\documentclass[11pt,a4paper]{article}
\usepackage[hyperref]{acl2017}
\usepackage{times}
\usepackage{latexsym}
\usepackage{amsmath}
\usepackage{amssymb}
\usepackage{url}
\usepackage{multirow}

\usepackage{array}
\newcolumntype{P}[1]{>{\centering\arraybackslash}p{#1}}
\usepackage{graphicx}
\usepackage{caption}
\usepackage[export]{adjustbox}
\usepackage{lipsum}

\aclfinalcopy % Uncomment this line for the final submission
\def\aclpaperid{3146} %  Enter the acl Paper ID here

\newcommand{\M}{\mathbf}

\DeclareMathOperator*{\argmax}{argmax}

\title{A Generative Parser with a Discriminative Recognition Algorithm}

\author{Jianpeng Cheng\quad Adam Lopez \and Mirella Lapata\\
	School of Informatics, University of Edinburgh\\
	{\tt jianpeng.cheng@ed.ac.uk, \{alopez,mlap\}@inf.ed.ac.uk}
}

\date{}

\begin{document}
\maketitle
\begin{abstract}
  Generative models defining joint distributions over parse trees and
  sentences are useful for parsing and language modeling, but impose
  restrictions on the scope of features and are often outperformed by
  discriminative models.  We propose a framework for parsing and
  language modeling which marries a generative model with a
  discriminative recognition model in an encoder-decoder setting. We
  provide interpretations of the framework based on expectation
  maximization and variational inference, and show that it enables
  parsing and language modeling within a single implementation. On the
  English Penn Treenbank, our framework obtains competitive
  performance on constituency parsing while matching the state-of-the-art single-model
  language modeling score.\footnote{Our code is available at
    \url{https://github.com/cheng6076/virnng.git}.}
\end{abstract}

\section{Introduction}
Generative models defining joint distributions over parse trees and sentences are good theoretical models for interpreting natural language data, and appealing tools for tasks such as parsing, grammar induction and language modeling \cite{Collins:1999,henderson2003,titov2007c,petrov2007improved,dyer2016recurrent}.
However, they often impose strong independence assumptions which restrict the use of arbitrary features for effective disambiguation.
Moreover, generative parsers are typically trained by maximizing the joint probability of the parse tree and the sentence---an objective that only indirectly relates to the goal of parsing. 
At test time, these models require a relatively expensive recognition algorithm \cite{Collins:1999,titov2007c} to recover the parse tree, but 
the parsing performance consistently lags behind their discriminative competitors \cite{nivre2007maltparser,huang2008forest,goldberg2010efficient}, which are directly trained to maximize the conditional probability of the parse tree given the sentence, where linear-time decoding algorithms exist (e.g.,~for transition-based parsers).

In this work, we propose a parsing and language modeling framework
that marries a generative model with a discriminative recognition
algorithm in order to have the best of both worlds.  The idea of
combining these two types of models is not new. For example,
\newcite{collins2005discriminative} propose to use a generative model
to generate candidate constituency trees and a discriminative model to
rank them.  \newcite{sangati2009generative} follow the opposite
direction and employ a generative model to re-rank the dependency
trees produced by a discriminative parser.  However, previous work
combines the two types of models in a goal-oriented, pipeline fashion,
which lacks model interpretations and focuses solely on parsing.

In comparison, our framework unifies generative and discriminative
parsers with a single objective, which connects to expectation
maximization and variational inference in grammar induction settings.
In a nutshell, we treat parse trees as latent factors generating
natural language sentences and parsing as a posterior inference task.
We showcase the framework using Recurrent Neural Network Grammars
(RNNGs; \citealt{dyer2016recurrent}), a recently proposed
probabilistic model of phrase-structure trees based on neural
transition systems.  Different from this work which introduces
separately trained discriminative and generative models, we integrate
the two in an auto-encoder which fits our training objective.  We show
how the framework enables grammar induction, parsing and language
modeling within a single implementation. On the English Penn Treebank,
we achieve competitive performance on constituency parsing and
state-of-the-art single-model language modeling score.

%single-sentence language modeling score 99.8 on English.

\section{Preliminaries \label{preliminary}}

In this section we briefly describe Recurrent Neural Network Grammars
(RNNGs; \citealt{dyer2016recurrent}), a top-down
transition-based algorithm for parsing and generation.  There are two
versions of RNNG, one discriminative, the other generative. We follow
the original paper in presenting the discriminative variant first. 

The discriminative RNNG follows a shift-reduce parser that converts a
sequence of words into a parse tree.  As in standard shift-reduce
parsers, the RNNG uses a buffer to store unprocessed terminal symbols
and a stack to store partially completed syntactic constituents.  At
each timestep, one of the following three operations\footnote{To be
  precise, the total number of operations under our description is
  $|$X$|$+2 since the \textsc{nt} operation varies with the
  non-terminal choice X.} is performed:
\begin{itemize}

\item \textsc{nt}(X) introduces an open non-terminal X onto the top of
  the stack, represented as an open parenthesis followed by X,
  e.g.,~(NP.

\item \textsc{shift} fetches the terminal in the front of the buffer
  and pushes it onto the top of the stack.

\item \textsc{reduce} completes a subtree by repeatedly popping the
  stack until an open non-terminal is encountered. The non-terminal is
  popped as well, after which a composite term representing the entire
  subtree is pushed back onto the top of the stack, e.g., (NP
  \textit{the cat}).
\end{itemize}

The above transition system can be adapted with minor modifications to
an algorithm that generates trees and sentences. In generator
transitions, there is no input buffer of unprocessed words but there
is an output buffer for storing words that have been generated. To
reflect the change, the previous \textsc{shift} operation is modified
into a \textsc{gen} operation defined as follows:
\begin{itemize}
	\item \textsc{gen} generates a terminal symbol and add it to the stack and the output buffer. 
\end{itemize}

\section{Methodology}
\label{sec:methodology}

Our framework unifies generative and discriminative parsers within a
single training objective.  For illustration, we adopt the two RNNG
variants introduced above with our customized features.  Our starting
point is the generative model (\S~\ref{generative model}), which
allows us to make explicit claims about the generative process of
natural language sentences. Since this model alone lacks a bottom-up
recognition mechanism, we introduce a discriminative recognition model
(\S~\ref{recognition model}) and connect it with the generative
model in an encoder-decoder setting.  To offer a clear interpretation
of the training objective (\S~\ref{training}), we first consider
the parse tree as latent and the sentence as observed. We then discuss
extensions that account for labeled parse trees. Finally, we present
various inference techniques for parsing and language modeling within
the framework (\S~\ref{inference}).

\subsection{Decoder (Generative Model) \label{generative model}}
The decoder is a generative RNNG that models the joint
probability~$p(x, y)$ of a latent parse tree~$y$ and an observed
sentence~$x$.  Since the parse tree is defined by a sequence of
transition actions~$a$, we write~$p(x, y)$ as~$p(x, a)$.\footnote{We
  assume that the action probability does not take the actual terminal
  choice into account.}  The joint distribution~$p(x, a)$ is
factorized into a sequence of transition probabilities and terminal
probabilities (when actions are \textsc{gen}), which are parametrized
by a transitional state embedding $\M{u}$:
\begin{eqnarray}
p(x, a) & = & p(a)p(x|a) \nonumber\\
        & = &\prod_{t=1}^{|a|} p(a_t | \M{u}_t) p(x_t | \M{u}_t)^{\mathbb{I}(a_t = \textsc{gen})} 
\end{eqnarray}
where $\mathbb{I}$ is an indicator function and $\M{u}_t$ represents the state embedding at time step $t$. 
Specifically, the conditional probability of the next action is:
\begin{equation}
p(a_t | \M{u}_t) = \frac{\exp (\M{a}_t \M{u}_t^\text{T} + b_a) }{\sum_{a' \in \mathcal{A}} \exp (\M{a}' \M{u}_t^\text{T} + b_{a'}) }
\label{nexta}
\end{equation}
where $\M{a}_t$ represents the action embedding at time step $t$,
$\mathcal{A}$ the action space and $b_{a}$ the bias.  Similarly, the
next word probability (when \textsc{gen} is invoked) is computed as:
\begin{equation}
p(w_t | \M{u}_t) =  \frac{\exp (\M{w}_t \M{u}_t^\text{T} + b_w) }{\sum_{w' \in \mathcal{W}} \exp (\M{w}' \M{u}_t^\text{T} + b_{w'}) }
\end{equation}
where $\mathcal{W}$ denotes all words in the vocabulary.

To satisfy the independence assumptions imposed by the generative
model, $\M{u}_t$ uses only a restricted set of features defined over
the output buffer and the stack --- we consider~$p(a)$ as a context
insensitive prior distribution.  Specifically, we use the following
features: 1) the stack embedding~$\M{d}_t$ which encodes the stack of
the decoder and is obtained with a stack-LSTM
\cite{dyer-EtAl:2015:ACL-IJCNLP,dyer2016recurrent}; 2) the output
buffer embedding $\M{o}_t$; we use a standard LSTM to compose the
output buffer and $\M{o}_t$ is represented as the most recent state of
the LSTM; and 3) the parent non-terminal embedding~$\M{n}_t$ which is
accessible in the generative model because the RNNG employs a depth-first generation
order. Finally, $\M{u}_t$ is computed as:
\begin{equation}
\M{u}_t = \M{W}_2 \tanh ( \M{W}_1 [\M{d}_t, \M{o}_t, \M{n}_t]  + b_d)
\end{equation}
where $\M{W}$s are weight parameters and $b_d$ the bias.

\subsection{Encoder (Recognition Model) \label{recognition model}}
The encoder is a discriminative RNNG that computes the conditional probability $q(a | x)$ of the transition action sequence $a$ given an observed sentence $x$.
This conditional probability is factorized over time steps as:
\begin{equation}
q(a | x) = \prod_{t=1}^{|a|} q(a_t | \M{v}_t)
\end{equation}
where $\M{v}_t$ is the transitional state embedding of the encoder at
time step~$t$.

The next action is predicted similarly to Equation~\eqref{nexta}, but
conditioned on $\M{v}_t$. Thanks to the discriminative property,
$\M{v}_t$ has access to any contextual features defined over the
entire sentence and the stack --- $q(a | x)$~acts as a context
sensitive posterior approximation. Our features\footnote{Compared to
  \newcite{dyer2016recurrent}, the new features we introduce are 3)
  and 4), which we found empirically useful.} are: 1)~the stack
embedding $\M{e}_t$ obtained with a stack-LSTM that encodes the stack
of the encoder; 2)~the input buffer embedding~$\M{i}_t$; we use a
bidirectional LSTM to compose the input buffer and represent each word
as a concatenation of forward and backward LSTM states; $\M{i}_t$ is
the representation of the word on top of the buffer; 3) to incorporate
more global features and a more sophisticated look-ahead mechanism for
the buffer, we also use an adaptive buffer embedding $\M{\bar{i}}_t$;
the latter is computed by having the stack embedding $\M{e}_t$ attend
to all remaining embeddings on the buffer with the attention function
in \newcite{vinyals2015grammar}; and 4)~the parent non-terminal
embedding $\M{n}_t$. Finally, $\M{v}_t$ is computed as follows:
\begin{equation}
\M{v}_t = \M{W}_4 \tanh ( \M{W}_3 [\M{e}_t, \M{i}_t, \M{\bar{i}}_t, \M{n}_t]  + b_e)
\end{equation}
where $\M{W}$s are weight parameters and $b_e$ the bias.

\subsection{Training \label{training}}
Consider an auto-encoder whose encoder infers the latent parse tree
and the decoder generates the observed sentence from the parse
tree.\footnote{Here, \textsc{gen} and \textsc{shift} refer to the same
  action with different definitions for encoding and decoding.}  The
maximum likelihood estimate of the decoder parameters is determined by
the log marginal likelihood of the sentence:
\begin{equation}
\log p(x) = \log \sum_a p(x, a)
\end{equation}
We follow expectation-maximization and variational inference
techniques to construct an evidence lower bound of the above quantity
(by Jensen's Inequality), denoted as follows:
\begin{equation}
\log p(x) \geq \mathbb{E}_{q(a|x)} \log \frac{p(x, a)}{q(a|x)} = \mathcal{L}_x
\label{obj1}
\end{equation}
where~$p(x,a) = p(x | a)p(a)$ comes from the decoder or the generative
model, and~$q(a|x)$ comes from the encoder or the recognition model.
The objective function\footnote{See \S~\ref{rw} and Appendix
  \ref{sec:supplemental} for comparison between this objective and the
  importance sampler of \newcite{dyer2016recurrent}.} in
Equation~\eqref{obj1}, denoted by $\mathcal{L}_x$, is unsupervised and
suited to a grammar induction task. This objective can be optimized with the methods shown in \newcite{miao2016language}.

Next, consider the case when the parse tree is observed. We can
directly maximize the log likelihood of the parse tree for the encoder
output~$\log q(a|x) $ and the decoder output~$\log p(a)$:
\begin{equation}
\mathcal{L}_a= \log q(a|x) + \log p(a)
\label{obj2}
\end{equation}
This supervised objective leverages extra information of labeled parse trees to
regularize the distribution $q(a|x)$ and $p(a)$, and the final objective is: 
\begin{equation}
\mathcal{L} =  \mathcal{L}_x + \mathcal{L}_a
\label{obj}
\end{equation}
where $\mathcal{L}_x$ and $\mathcal{L}_a$ can be balanced with the task focus (e.g, language modeling or parsing).

\subsection{Inference \label{inference}}
We consider two inference tasks, namely parsing and language modeling.

\paragraph{Parsing} In parsing, we are interested in the parse tree
that maximizes the posterior $p(a|x)$ (or the joint $p(a, x)$).
However, the decoder alone does not have a bottom-up recognition
mechanism for computing the posterior. Thanks to the
encoder, we can compute an approximated
posterior~$q(a|x)$ in linear time and select the parse tree that maximizes this
approximation.  An alternative is to generate candidate trees by
sampling from~$q(a|x)$, re-rank them with respect to the
joint~$p(x, a)$ (which is proportional to the true posterior), and
select the sample that maximizes the true posterior. 

\paragraph{Language Modeling} In language modeling, our goal is to
compute the marginal probability \mbox{$p(x) = \sum_a p(x, a)$}, which
is typically intractable.  To approximate this quantity, we can use
Equation~\eqref{obj1} to compute a lower bound of the log likelihood~$\log p(x)$ and then exponentiate it to get a pessimistic
approximation of~$p(x)$.\footnote{As a reminder, the language modeling objective is $\exp(NLL/T)$, where $NLL$ denotes the total negative log likelihood of the test data and $T$ the token counts.} 

Another way of computing $p(x)$ (without lower bounding) would be to
use the variational approximation~$q(a|x)$ as the proposal
distribution as in the importance sampler of
\newcite{dyer2016recurrent}. We discuss details in
Appendix~\ref*{sec:supplemental}.

\begin {table}[t]
\begin{center}
%	\small
	\begin{tabular}{l | r}
		\hline
		Learned word embedding dimensions  & 40 \\
		Pretrained word embedding dimensions & 50 \\
		POS tag embedding dimensions & 20 \\
		Encoder LSTM dimensions & 128 \\
		Decoder LSTM dimensions & 256 \\
		LSTM layer & 2 \\
		Encoder dropout & 0.2 \\
		Decoder dropout & 0.3 \\
		\hline
	\end{tabular}
\end{center}
%\vspace{-2ex} 
\caption{Hyperparameters.}
\label{hyper}
\vspace{-2ex}
\end{table}

\section{Related Work \label{rw}}
Our framework is related to a class of variational autoencoders
\cite{kingma2013auto}, which use neural networks for posterior
approximation in variational inference.  This technique has been
previously used for topic modeling \cite{miao2016neural} and sentence
compression \cite{miao2016language}.  Another interpretation of the
proposed framework is from the perspective of guided policy search in
reinforcement learning \cite{bachman2015data}, where a
generative parser is trained to imitate the trace of a discriminative parser.
Further connections can be drawn with the importance-sampling based
inference of \newcite{dyer2016recurrent}.  There, a generative RNNG
and a discriminative RNNG are trained separately; during language
modeling, the output of the discriminative model serves as the
proposal distribution of an importance sampler
$p(x) = \mathbb{E}_{q(a|x)} \frac{p(x,a)}{q(a|x)}$.  Compared to their
work, we unify the generative and discriminative RNNGs in a single
framework, and adopt a joint training objective.

\section{Experiments}
\label{sec:experiments}

\begin {table}[t]
\begin{center}
%	\small
	\begin{tabular}{ p{2cm} | l | r }
		\hline
	\multirow{ 5}{*}{\parbox{2cm}{\hspace*{0ex}Discriminative parsers}} & \newcite{socher2013parsing}  & 90.4 \\
	&	\newcite{zhu2013fast} & 90.4 \\
	&	\newcite{dyer2016recurrent}   & 91.7 \\
	&	\newcite{cross2016incremental}   & 89.9 \\
	&	\newcite{vinyals2015grammar}  &  92.8 \\
    	\hline
	\multirow{ 3}{*}{\parbox{2cm}{Generative parsers}} &	\newcite{petrov2007improved}  & 90.1 \\
	&	\newcite{shindo2012bayesian}  & 92.4 \\
	&	\newcite{dyer2016recurrent} & 93.3 \\
		\hline
	\multirow{ 2}{*}{This work} &	$\argmax_a q(a|x)$ & 89.3  \\
	&	$\argmax_a p(a,x)$ & 90.1   \\ 
		\hline
	\end{tabular}
\end{center}
%\vspace{-2ex} 
\caption{Parsing results (F1) on the PTB test set. }
\label{parsing}
\vspace{-2ex}
\end{table}

We performed experiments on the English Penn Treebank dataset; we used
sections~2--21 for training,~24 for validation, and~23
for testing.  Following \newcite{dyer-EtAl:2015:ACL-IJCNLP}, we
represent each word in three ways: as a learned vector, a
pretrained vector, and a
POS tag vector.  The encoder word embedding is the concatenation of
all three vectors while the decoder uses only the first two since we
do not consider POS tags in generation.  Table~\ref{hyper} presents
details on the hyper-parameters we used.  To find the MAP parse tree
$\argmax_a p(a, x)$ (where $p(a, x)$ is used rank the output of
$q(a|x)$) and to compute the language modeling perplexity (where $a \sim q(a|x)$), we collect 100 samples
from $q(a|x)$, same as \newcite{dyer2016recurrent}.

Experimental results for constituency parsing and language modeling
are shown in Tables~\ref{parsing} and~\ref{lm}, respectively.  As can
be seen, the single framework we propose obtains competitive parsing
performance. Comparing the two inference methods for parsing, ranking approximated MAP trees from $q(a|x)$ with respect
to $p(a, x)$ yields a small improvement, as in \newcite{dyer2016recurrent}.  It is worth noting that our
parsing performance lags behind \newcite{dyer2016recurrent}. We
believe this is due to implementation disparities, such as the
modeling of the reduce operation. While \newcite{dyer2016recurrent}
use an LSTM as the syntactic composition function of each subtree, we
adopt a rather simple composition function based on
embedding averaging, which gains computational efficiency but loses accuracy. 

On language modeling, our framework achieves lower perplexity compared
to \newcite{dyer2016recurrent} and baseline models.
This gain possibly comes from the joint optimization of both the generative 
and discriminative components towards a language modeling objective.
However, we acknowledge a subtle difference between
\newcite{dyer2016recurrent} and our approach compared to baseline
language models: while the latter incrementally estimate the next word
probability, our approach (and \citealt{dyer2016recurrent}) directly
assigns probability to the entire sentence.
Overall, the advantage of our framework compared to \newcite{dyer2016recurrent} is that
it opens an avenue to unsupervised training.

\begin {table}[t]
\begin{center}
%	\small
	\begin{tabular}{ l | r }
		\hline
		KN-5 & 255.2 \\
		LSTM & 113.4 \\
		\newcite{dyer2016recurrent} & 102.4 \\
		\hline
		This work: $a \sim q(a|x)$ & 99.8  \\ \hline
	\end{tabular}
\end{center}
%\vspace{-2ex} 
\caption{Language modeling results (perplexity). }
\label{lm}
\vspace{-2ex}
\end{table}

\section{Conclusions}
\label{sec:conclusions}

We proposed a framework that integrates a generative parser with a
discriminative recognition model and showed how it can be instantiated
with RNNGs.  We demonstrated that a unified framework, which relates
to expectation maximization and variational inference, enables
effective parsing and language modeling algorithms.  Evaluation on the
English Penn Treebank, revealed that our framework obtains competitive
performance on constituency parsing and state-of-the-art
results on single-model language modeling.  In the future, we would like to perform
grammar induction based on Equation~\eqref{obj1}, with gradient
descent and posterior regularization techniques
\cite{ganchev2010posterior}.

\paragraph{Acknowledgments} We thank three anonymous reviewers and members of the ILCC for valuable
feedback, and Muhua Zhu and James Cross for help with data preparation. The support of the European Research Council under award
number 681760 ``Translating Multiple Modalities into Text'' is
gratefully acknowledged.

% include your own bib file like this:
%\bibliographystyle{acl}
%\bibliography{acl2017}

\appendix

\section{Comparison to Importance Sampling \cite{dyer2016recurrent}}
\label{sec:supplemental}
In this appendix we highlight the connections between importance
sampling and variational inference, thereby comparing our method with
\newcite{dyer2016recurrent}.

Consider a simple directed graphical model with discrete latent
variables~$a$ (e.g., $a$ is the transition action sequence) and
observed variables~$x$ (e.g., $x$ is the sentence).  The model
evidence, or the marginal likelihood $p(x) = \sum_a p(x, a)$ is often
intractable to compute. Importance sampling transforms the above
quantity into an expectation over a distribution~$q(a)$, which is
known and easy to sample from:
\begin{equation}
p(x) = \sum_a p(x, a) \frac{q(a)}{q(a)} =  \mathbb{E}_{q(a)} w(x, a)
\label{is}
\end{equation}
where~$q(a)$ is the proposal distribution
and~$w(x, a) = \frac{p(x, a) }{q(a)}$ the importance weight.  The
proposal distribution can potentially depend on the observations~$x$,
i.e.,~$q(a) \triangleq q(a|x)$.

A challenge with importance sampling lies in choosing a proposal
distribution which leads to low variance.  As shown in
\newcite{rubinstein2008simulation}, the optimal choice of the proposal
distribution is in fact the true posterior~$p(a|x)$, in which case the
importance weight $\frac{p(a,x)}{p(a|x)} = p(x)$ is constant with
respect to~$a$.  In \newcite{dyer2016recurrent}, the proposal
distribution depends on~$x$, i.e.,~$q(a) \triangleq q(a|x)$, and is
computed with a separately-trained, discriminative model.  This
proposal choice is close to optimal, since in a fully supervised
setting $a$~is also observed and the discriminative model can be
trained to approximate the true posterior well.  We hypothesize that
the performance of their importance sampler is dependent on this
specific proposal distribution.  Besides, their training strategy does
not generalize to an unsupervised setting.

In comparison, variational inference approach approximates the log
marginal likelihood $\log p(x)$ with the evidence lower bound. It is
a natural choice when one aims to optimize Equation~\eqref{is}
directly:
\begin{equation}
\begin{split}
 \log p(x) &  = \log \sum_a p(x, a) \frac{q(a)}{q(a)} \\
& \geq \mathbb{E}_{q(a)} \log \frac{ p(x, a)}{q(a)}
\label{elbotrain}
 \end{split}
\end{equation}
where $q(a)$ is the variational approximation of the true
posterior. Again, the variational approximation can potentially depend
on the observation~$x$ (i.e., $q(a) \triangleq q(a|x)$) and can be
computed with a discriminative model.  Equation~\eqref{elbotrain} is a
well-defined, unsupervised training objective which allows us to
jointly optimize generative (i.e., $p(x, a)$) and discriminative
(i.e., $q(a|x)$) models.  To further support the observed
variable~$a$, we augment this objective with supervised terms shown
in Equation~\eqref{obj}, following \newcite{kingma2014semi} and
\newcite{miao2016language}.

Equation~\eqref{elbotrain} can be also used to approximate the
marginal likelihood~$p(x)$ (e.g., in language modeling) with its lower
bound.  An alternative choice without lower bounding is to use the
variational approximation $q(a|x)$ as the proposal distribution in
importance sampling (Equation~\eqref{is}).
\newcite{ghahramani1999variational} show that this proposal
distribution leads to improved results of importance samplers.

\bibliography{acl2017}
\bibliographystyle{acl_natbib}

\end{document}